\documentclass[twoside,11pt]{article}

\usepackage{blindtext}

\usepackage{cite}
\usepackage{hyperref}
\usepackage{jmlr2e}
\usepackage{minted}
\usepackage{amsmath}
\usepackage{booktabs}
\usepackage{multirow}
\usepackage{listings}
\usepackage{xcolor}
\usepackage{newfloat}
\usepackage{caption}

\DeclareFloatingEnvironment[
  name=Code,
  listname={List of Codes},
  fileext=loc
]{code}

\usepackage{svg}
\usepackage{float}
\floatname{listing}{Code}

\usepackage{lastpage}

\ShortHeadings{3D Transport-based Morphometry (3D-TBM) for medical image analysis}{3D Transport-based Morphometry (3D-TBM) for medical image analysis}
\firstpageno{1}

\begin{document}

\title{3D Transport-based Morphometry (3D-TBM) for medical image analysis}

\author{\name Hongyu Kan \email tuy7xa@virginia.edu \\
       \addr Department of Computer Engineering\\
       University of Virginia\\
       Charlottesville, VA 22903, USA
       \AND
       \name Kristofor Pas \email xfd3sy@virginia.edu \\
       \addr Department of Biomedical Engineering\\
       University of Virginia\\
       Charlottesville, VA 22908, USA
       \AND
       \name Ivan Medri \email pzr7pr@virginia.edu \\
       \addr Department of Biomedical Engineering\\
       University of Virginia\\
       Charlottesville, VA 22908, USA
       \AND
       \name Naqib Sad Pathan \email qpb3vt@virginia.edu \\
       \addr Department of Electrical and Computer Engineering\\
       University of Virginia\\
       Charlottesville, VA 22908, USA
       \AND
       \name Natasha Ironside \email ni8vb@virginia.edu \\
       \addr Department of Neurological Surgery\\
       University of Virginia\\
       Charlottesville, VA 22908, USA
       \AND
       \name Shinjini Kundu \email kundu@wustl.edu \\
       \addr Department of Radiology\\
       Washington University in St. Louis\\
       St. Louis, MO 63110, USA
       \AND
       \name Jingjia He \email h.jingjia@wustl.edu \\
       \addr Department of Electrical and Systems Engineering\\
       Washington University in St. Louis\\
       St. Louis, MO 63130, USA
       \AND
       \name Gustavo Kunde Rohde \email gustavo@virginia.edu \\
       \addr Department of Biomedical Engineering and Electrical and Computer Engineering\\
       University of Virginia\\
       Charlottesville, VA 22908, USA}
       
\editor{My editor}

\maketitle
\begin{abstract}
Transport-Based Morphometry (TBM) has emerged as a new framework for 3D medical image analysis. By embedding images into a transport domain via invertible transformations, TBM facilitates effective classification, regression, and other tasks using transport-domain features. Crucially, the inverse mapping enables the projection of analytic results back into the original image space, allowing researchers to directly interpret clinical features associated with model outputs in a spatially meaningful way. To facilitate broader adoption of TBM in clinical imaging research, we present 3D-TBM, a tool designed for morphological analysis of 3D medical images. The framework includes data preprocessing, computation of optimal transport embeddings, and analytical methods such as visualization of main transport directions, together with techniques for discerning discriminating directions and related analysis methods. We also provide comprehensive documentation and practical tutorials to support researchers interested in applying 3D-TBM in their own medical imaging studies. The source code is publicly available through PyTransKit \cite{pytranskit2023}.

\end{abstract}


\section{Overview of TBM}

Three-dimensional (3D) transport-based morphometry (TBM) \cite{kundu2018discovery, wang2013linear} has recently emerged as a promising framework for morphological analysis in medical imaging, with applications including the prediction of hematoma expansion\cite{ironside2024fully}, the study of brain tissue distribution changes associated with cardiorespiratory fitness\cite{kundu2021investigating}, and the identification of autism-related endophenotypes\cite{kundu2024discovering}. The key idea is based on the mathematical principles of optimal mass transport \cite{villani2008optimal}, which embed medical images into a novel mathematical space called the Linear Optimal Transport (LOT) \cite{wang2013linear} space. In the transport domain, morphological variations can be captured and analyzed in a more direct and linear manner \cite{kundu2024discovering}. Moreover, since the optimal transport transformation is invertible, the analysis results can be 'mapped back' to the original image space. This property not only enables classification and regression tasks, but also allows visualization of anatomical regions in the original images that are associated with the model outputs, thus providing clinical interpretability.

Despite the significant advantages of 3D TBM in medical image analysis there is currently a lack of user-friendly software readily available for clinical researchers. With the growing interest in optimal transport within clinical research, and medical imaging in particular, an increasing number of researchers are facing difficulties due to the lack of suitable tools. Therefore, providing a user-friendly and accessible framework for 3D TBM has become particularly important.

In this work, we present 3D-TBM, a Python-based tool to enable TBM to be applied to 3D medical imaging. Its algorithm is based on the work of \cite{kundu2018discovery} and provides a complete pipeline covering data preprocessing, computing the linear optimal transport (LOT) embedding \cite{wang2013linear, kolouri2017optimal, basu2014detecting}, and data analysis. The corresponding components include preprocessing for the calculation of LOT, the LOT computation process \cite{kundu2018discovery}, principal component analysis (PCA) \cite{jolliffe2016principal}, linear discriminant analysis (LDA/PLDA) \cite{balakrishnama1998linear,wang2011penalized}, canonical correlation analysis (CCA) \cite{hardoon2004canonical}, and visualization of analytical results. In addition, 3D-TBM offers a simple tutorial based on a small data set, allowing users to become familiar with the tool from the ground up. Each module is encapsulated within a dedicated interface and users can flexibly adjust the parameter settings according to their specific needs. The source code is publicly available through PyTransKit\cite{pytranskit2023}.



\section{Existing Tools for 3D medical imaging morphometry}

Although other medical imaging tools offer convenient functionalities for morphological analysis - such as registration, segmentation, and even neural network training - they do not provide tools to perform transport-based morphometry. Advanced Normalization Tools (ANTs) is a C++-based command-line library for high-dimensional biomedical image registration and analysis, enabling the statistical exploration, visualization, and integration of large-scale imaging data across modalities, species, and organ systems \cite{avants2009advanced}. Extensions of ANTs include ANTsR and ANTsPy, as well as ANTsX, which integrates deep learning methods into the ANTs framework \cite{tustison2021antsx,tustison2024antsx}. MONAI is an open-source, community-driven PyTorch-based framework that extends PyTorch for medical imaging, providing specialized architectures, transformations, and utilities to facilitate the development and deployment of healthcare AI models \cite{cardoso2022monai}.
FreeSurfer is a comprehensive neuroimaging analysis suite that quantifies functional, connectional, and structural brain properties, evolving from cortical surface modeling to the automated reconstruction of most macroscopically visible brain structures from T1-weighted images \cite{fischl2012freesurfer,dale1999cortical}. SPM is a software package for analyzing brain imaging data sequences, including cross-sectional datasets from multiple cohorts and longitudinal time-series from individual subjects \cite{friston1994statistical,penny2011statistical}. AFNI (Analysis of Functional NeuroImages) is a comprehensive software suite composed of C, Python, and R programs, along with shell scripts, designed primarily for analyzing and visualizing various MRI modalities, including anatomical, functional (fMRI), and diffusion-weighted (DW) data\cite{cox1996afni,cox1997software}.

However, to the best of our knowledge, these widely used tools do not support 3D TBM. Other libraries related to optimal transport, such as POT \cite{flamary2021pot}, do not have specific functions for TBM. Their use typically involves cumbersome data preprocessing followed by the reconstruction of custom scripts to invoke the desired functionalities, making the workflow both complex and time-consuming. To overcome these challenges, we present 3D-TBM, a TBM tool specifically designed for 3D medical imaging that supports LOT computation, model analysis, and result visualization, thereby enabling classification and regression tasks on 3D image data.

\section{Transport-based Morphometry (TBM)}
\begin{figure}[htbp]
    \centering
    \includegraphics[width=1\linewidth]{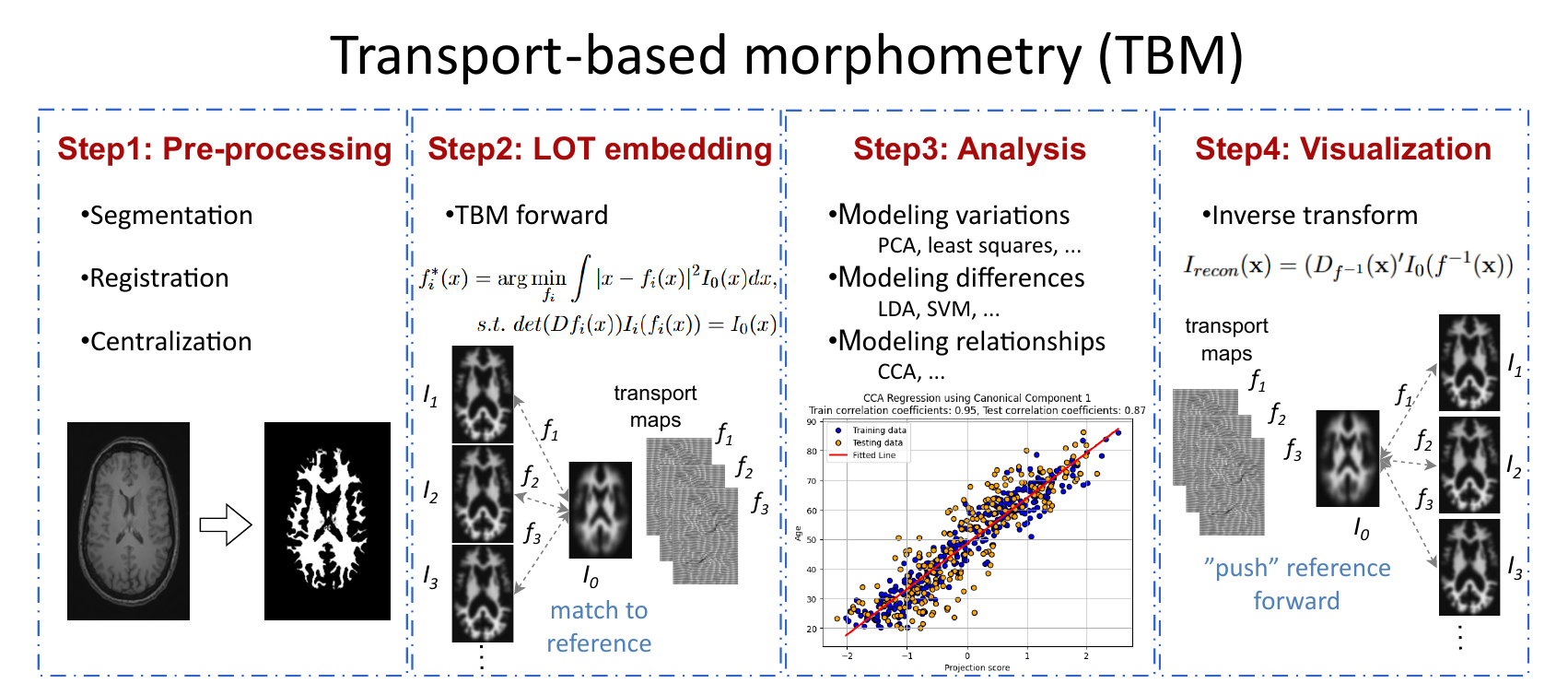}
    \caption{The workflow of Transport-Based Morphometry (TBM). It includes 4 steps: step 1: preparing data for 3D-TBM; step 2: embedding 3D images into the transport domain; step 3: using the model for variable analysis, classification, and regression; and step 4: visualizing the analysis results.}
    \label{fig:placeholder}
\end{figure}

In this section, we present the 3D-TBM in detail, a Python-based tool that encapsulates complex mathematical logic and computations into simple interfaces, with the aim of providing researchers interested in TBM with a convenient and accessible framework. 3D-TBM consists of three steps: embedding 3D images into the transport domain; using the model for variable analysis, classification, and regression; and visualizing the analysis results. As shown in Figure \ref{fig:placeholder}. 

To better demonstrate the functionality of 3D-TBM, we employed the IXI dataset\cite{ixi_dataset} as an example and conducted several illustrative experiments on it. The IXI dataset provides a publicly accessible set of multi-modal brain MRI scans acquired across multiple imaging centers, including T1, T2, and PD images. For demonstration purposes, we investigated the chronological age as reflected in the white matter of the participants. Specifically, we investigated whether a given sample could be classified as a young adult (under 35 years of age) or an older adult (over 60 years of age), or whether age could be quantitatively predicted through regression based on white matter features. 

\subsection{Step 1: Pre-processing \& preparation for TBM}


All 3D images must undergo preprocessing before being used with 3D-TBM. This pre-processing typically involves registration, segmentation, cropping, and centering, with specific procedures varying according to the task. For example, in the prediction of hematoma expansion, the hematoma regions must be segmented and cropped \cite{ironside2024fully}, whereas in white matter classification tasks, the white matter regions should be segmented \cite{kundu2024discovering}, \cite{kundu2021investigating}. 
Since these preprocessing steps depend on the anatomical regions of interest and the particular application, 3D-TBM does not provide built-in implementations for them, leaving this responsibility to the user. However, it is important to note that centering is a mandatory step to ensure that all processed images are aligned around a consistent center.

For ease of data loading, all preprocessed images should be stored as NumPy arrays \cite{harris2020array}. We assume that the data set includes a set of 3D medical images $I_1,...,I_N: R^{h\times w \times d} \to R^+$, where $h,w, d$ is the dimension of the 3D medical images and $N$ is the number of 3D medical images. The training and testing sets have already been partitioned, let $I_1,...I_K$ belong to the training set and $I_{K+1},...I_N$ belong to the testing set, where $1<K<N$, with all file paths and their corresponding labels organized in CSV files, as shown in Table 1, where the 'image\_path' can be read directly, 'label' is the corresponding label. You can organize other variables related to the data in this file, but we don’t need them here.

\begin{table}[h]
\centering
\caption{Example of the CSV file}
\begin{tabular}{lcc}
\toprule
image\_path & label & other variable... \\ 
\midrule
./data/0.npy  & 1 & ... \\
./data/1.npy  & 0 & ... \\
... & ... & ... \\
\bottomrule
\end{tabular}
\label{tab:performance}
\end{table}

In the IXI data set samples employed in this study, the preprocessing pipeline consisted of spatial registration of all images, segmentation to extract white matter regions, and subsequent centering. All preprocessing procedures were performed using SPM and the resulting data were converted to NumPy format for further analysis.

The corresponding label data for these classification and regression tasks were organized into separate CSV files. The fully preprocessed dataset will be made publicly available to support future research and reproducibility.

\subsection{Step 2: Computing the transport embedding}


Before computing the LOT embedding, all images must be normalized according to Equation (\ref{eq:noraml}) \cite{kundu2018discovery}.


\begin{equation}
    \sum_x I_i(\textbf{x}) = 1, \ i= 1,...,N
\label{eq:noraml}
\end{equation}

This normalization can either be performed by the user during dataset preparation or, if omitted, will be automatically handled by the preprocessing module of 3D-TBM when loading the data. In addition, a reference image is required to compute the optimal transport maps. This reference can be specified by the user; otherwise, the mean of all training images is computed and used as the reference, as shown in Equation (\ref{eq:image mean}). 

\begin{equation}
    I_0(\textbf{x}) = \frac{1}{K}\sum_{i=1}^K I_i(\textbf{x})
\label{eq:image mean}
\end{equation}

\begin{figure}[!h]
    \centering
    \includegraphics[width=0.3\linewidth]{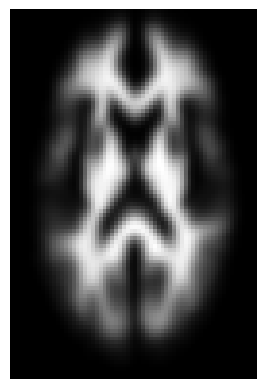}
    \caption{Intrinsic mean}
    \label{fig:intrinsic}
\end{figure}

In addition, using a better reference can further improve the predictive performance of the classifier. For example, the intrinsic mean, also referred to as the Wasserstein barycenter, can be iteratively updated using the mean optimal transport map, a process that progressively sharpens the template and yields a more representative intrinsic mean image for the data set \cite{kolouri2016continuous, agueh2011barycenters}. However, computing the intrinsic mean requires both the LOT embedding and the inverse visualization (introduced in subsequent sections), and its formulation is given in Equation (\ref{eq:intrinsic}). We defer the detailed discussion of this procedure to later sections.

\begin{equation}
    I_{mean} =  (\bar{f}^{-1})'I_0\circ \bar{f}^{-1},\ \  \bar{f}=\frac{1}{K}\sum_{i=1}^K f_i
\label{eq:intrinsic}
\end{equation}
Figure \ref{fig:intrinsic} presents an intrinsic mean of the IXI data set, while Code \ref{code:im} provides the technique for computing the intrinsic mean. However, this computation requires the LOT embedding to be completed beforehand (see subsequent sections).

\begin{code}[t]
\caption{Obtain the intrinsic mean (Figure~\ref{fig:intrinsic}).}
\label{code:im}
\begin{lstlisting}[language=Python]
from pytranskit.TBM3D.subspace import *
train_features, train_labels = ns_features(train_results)
Im = intrinsic_mean_ns(np.mean(train_features, axis=0), reference)
\end{lstlisting}
\end{code}



Code \ref{code:loading} provides the data-loading procedure and, when no reference image is specified, returns the mean image (Equation (\ref{eq:image mean})) as the default reference.
\begin{code}[h]
\caption{The code for loading data and calculating the mean image.}
\label{code:loading}
\begin{lstlisting}[language=Python]
from pytranskit.TBM3D.image_processing import load_data
train_file_path = './IXI_data/train.csv' #the csv path to training data
test_file_path = './IXI_data/test.csv' #the csv path to testing data
reference_path = None #the path to the reference image
normalized = False #if users already normalized all the images

train_images, train_labels, reference = load_data(file_path=train_file_path, 
        reference_path=reference_path, normalized=normalized, mode = 'train')
test_images, test_labels = load_data(file_path=test_file_path, 
        normalized=normalized, mode = 'test')
\end{lstlisting}
\end{code}

Upon completion of data loading, all images and labels are returned in separate lists. In the case of training data, an additional reference is also provided for use in the subsequent computation of the LOT embedding.



The linear optimal transport (LOT) framework provides a mathematical embedding that linearizes variations between probability distributions, thereby facilitating efficient analysis of complex data such as medical images. LOT treats medical images as smooth density functions and computes an optimal transport map that transfers the mass of a reference image to match that of a target image under the minimal transportation cost. Formally, it solves the following mathematical problem \cite{kundu2018discovery}:

\begin{equation}
    \begin{aligned}
    f^*_i(\textbf{x}) = \arg \min_{f_i} \int |\textbf{x}-f_i(\textbf{x})|^2I_0(\textbf{x})d\textbf{x},\\ s.t.\ det(Df_i(\textbf{x}))
    I_i(f_i(\textbf{x})) = I_0(\textbf{x})
    \end{aligned}
\end{equation}
where $D$ denotes the Jacobian matrix, $f_i : R^3 \to R^3$ is a mass preserving mapping from $I_0$ to $I_i$. Since the transport map is invertible, it can also be regarded as a representation of medical images.



\begin{code}[h]
\caption{The code for LOT embedding}
\label{code:embeding}
\begin{lstlisting}[language=Python]
from pytranskit.TBM3D.TBM_forward import TBM3D_forward, process_features
#the output path if you want to save the maps
train_maps_folder = './IXI_data/train_maps' 
test_maps_folder = './IXI_data/test_maps'
train_results = TBM3D_forward(train_images, train_labels, reference, 
        output_folder = train_maps_folder, parallel = True, num_workers = 12)
train_features, train_labels = process_features(train_results,reference)
test_results = TBM3D_forward(test_images, test_labels, reference, 
        output_folder = test_maps_folder, parallel = True, num_workers = 12)
test_features, test_labels = process_features(test_results,reference)
\end{lstlisting}
\end{code}

The optimization is performed using a multiscale accelerated gradient descent method, with the computation encapsulated within a dedicated interface. For relatively large 3D medical images, the process can be time-consuming. To mitigate this, we provide a parallel computing option that allows users to reduce runtime by enabling parallel execution (parallel=TRUE) and specifying the number of threads (num\_workers), as illustrated in the code \ref{code:embeding}.

The results of the computation can be saved to a specified path, thereby avoiding redundant calculations in subsequent use. Additionally, to facilitate the analysis in Step 2, $process\_features()$ will convert the transport maps into a $N*D$ NumPy array using the application of $(f_i^*-I_d)*\sqrt{I_0}$, where $f_i^*$ is the transport map, $I_d$ is the identity function, $I_0$ is the reference, $D$ is the size of a transport map.

\begin{figure}[h]   
    \centering
    \includegraphics[width=0.85\linewidth]{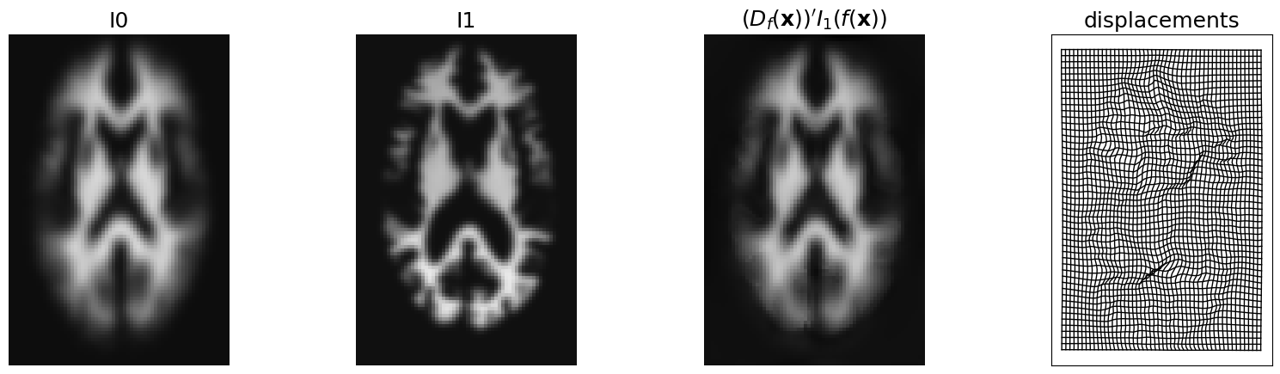}
    \caption{An example, through the transport map, the mass in the image is transported and aligned with the reference.}
    \label{fig:visual}
\end{figure}

\begin{figure}[h]   
    \centering
    \includegraphics[width=0.95\linewidth]{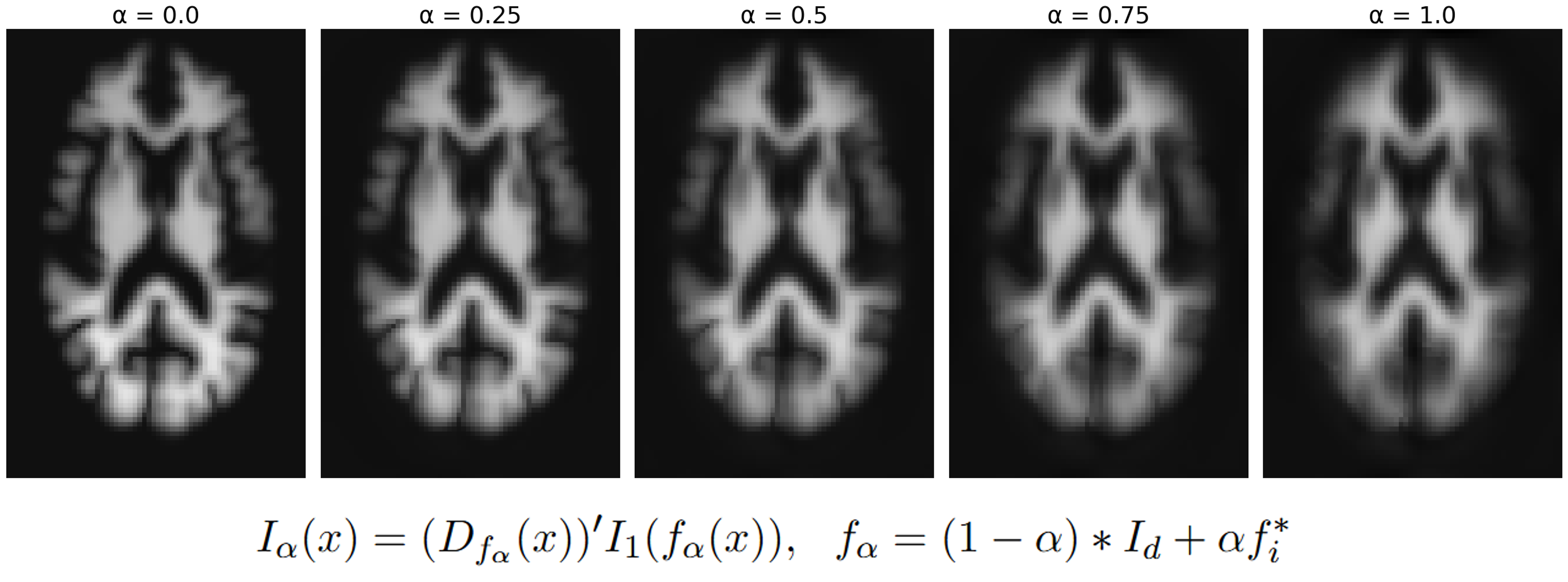}
    \caption{Visualize the geodesic from from $I_1$ to $I_0$}
    \label{fig:visual2}
\end{figure}

\begin{code}[h]
\caption{Visualize the embedding results(Figure \ref{fig:visual} \& Figure \ref{fig:visual2})}
\label{code:visual}
\begin{lstlisting}[language=Python]
from pytranskit.TBM3D.visualization import *
visualize_result(train_results[1])
\end{lstlisting}
\end{code}

3D-TBM includes a visualization interface that allows users to inspect the results of each computation. In this example, as illustrated in Figure \ref{fig:visual}, the mass in the image is transported and aligned with the reference, where $I_0$ denotes the reference, $I_1$ represents the input image, $(D_{f^{-1}}(\mathbf{x})'I_0( f^{-1}(\mathbf{x}))$ is the reconstruction of $I_0$ obtained from $I_1$ using the transport map, where $f^{-1}$ denotes the inverse transport map, and $displacements$ illustrates how the images are deformed.


This line of code also visualizes the geodesic from $I_1$ to $I_0$ through $f_{\alpha} = (1- \alpha)* I_d + \alpha f_i^*$, where the parameter $\alpha$ controls the linear combination between the identity map and the optimal transport map, as shown in Figure \ref{fig:visual2}. In this example, $\alpha = 0$ corresponds to the image $I_1$
, $\alpha = 1$ corresponds to $I_0$, and intermediate values yield the images along the geodesic.



\subsection{Step 3 \& 4: Statistical Analysis \& Visualization}


The features obtained from the LOT embedding can be utilized for model-based data analysis, including, but not limited to, dimensionality reduction, classification, and regression. Correspondingly, the principal components in PCA or the discriminant direction learned by PLDA can be mapped back to the original feature space, which are the optimal transport map features. According to Equation (\ref{eq:inverse}), the optimal transport map of an image can be inverted from the transport domain back to the image domain, thus reconstructing the 3D image. This allows us to directly examine the learned data representations in the image domain, thereby enhancing the interpretability of the models:

\begin{equation}
\label{eq:inverse}
    I_{recon}(\mathbf{x}) =  (D_{f^{-1}}(\mathbf{x})'I_0( f^{-1}(\mathbf{x}))
\end{equation}
where $f^{-1}$ stands for the inverse of deformation field $f$ and $D_{f^{-1}}(\mathbf{x})$ stands for the Jacobian determinant of the inverse deformation field at $\mathbf{x}$.

\subsubsection{Visualizing main transport modes with PCA}

Since the optimal transport map $f_i$ of a 3D medical image is a mapping from $R^3 \to R^3$, each voxel is associated with a three-dimensional vector. For an image of size $h \times w \times d$, the corresponding $f_i$ has a dimensionality of $3 * h * w * d$, which is typically very large. Directly analyzing such high-dimensional data is challenging; therefore, dimensionality reduction is a necessary step before any downstream analysis. 3D-TBM employs Principal Component Analysis (PCA) \cite{jolliffe2016principal}
 for this purpose. Suppose $\bar{f} = \frac{1}{K}\sum_{j=1}^{K}f_j$, and $\bar{f_i} = f_i - \bar{f}$, PCA is solving the following problem:

\begin{equation}
\label{eq:pca}
    \min_B \frac{1}{K}\sum_{i=1}^{K} \| BB^T\bar{f_i} - \bar{f_i} \|_2
\end{equation}
where $B=[\phi_1,\phi_2,...,\phi_k],\  \phi_i \in R^{3 * h * w * d},\  i = 1,...,k,\  k \ll 3 * h * w * d$, and then
\begin{equation}
\label{eq:pca_proj}
    x_i = B^T\bar{f_i} \in R^k
\end{equation}
becomes the representation of $f_i$, thus this approach enables the reduction of the dimension of the data. Code \ref{code:pca} provides an implementation:

\begin{code}[h]
\caption{The code for PCA}
\label{code:pca}
\begin{lstlisting}[language=Python]
from pytranskit.TBM3D.pca_tbm3d import SimplePCA
# train_features, test_features obtained from process_features()
# n_components: precentage of explained variance
pca = SimplePCA(n_components=0.96)
X_train_pca = pca.fit_transform(train_features)
X_test_pca = pca.transform(test_features)
\end{lstlisting}
\end{code}

This procedure eliminates dimensions that contribute minimally to the overall variance of the data set while retaining principal components that together explain 96\% of the variance (with this threshold adjustable) \cite{kundu2024discovering}.

\begin{figure}[!htbp]
    \centering
    \includegraphics[width=0.85\linewidth]{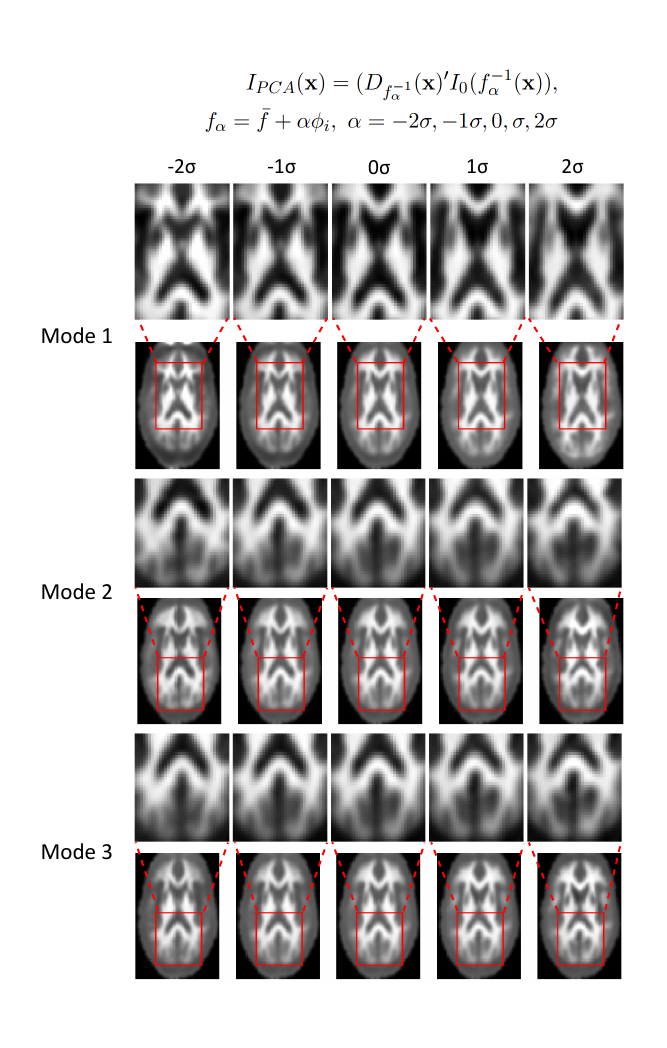}
    \caption{Image variations along the first three principal component directions derived from PCA.}
    \label{fig:pcamode}
\end{figure}

Visualizing the results of PCA helps to observe the distribution and characteristics of the data. The component vectors $\{\phi_1,\phi_2,...,\phi_k\}$ obtained from Equation (\ref{eq:pca}) capture the most representative directions of the data distribution. By sampling data points along these directions and mapping them back to the transport domain, the distributional characteristics of the data set can be visualized in an intuitive and interpretable manner(Equation (\ref{eq:pcavisual})).
\begin{equation}
    \begin{aligned}
        I_{PCA}(\mathbf{x})=(D_{f_\alpha^{-1}}(\mathbf{x})'I_0( f_\alpha^{-1}(\mathbf{x})),\\  f_\alpha = \bar{f}+\alpha \phi_i,\  \alpha = -2\sigma,-1\sigma,0,\sigma,2\sigma
    \end{aligned}
\label{eq:pcavisual}     
\end{equation}
where $\sigma$ is the standard deviation of PCA loadings for component vectors $\phi_i$. Figure \ref{fig:pcamode} illustrates the variations of the images along the first three principal components obtained by PCA according to Equation (\ref{eq:pcavisual}).



Code \ref{code:pca_viusal} can be used to observe the projection of data on the first 2 PCA directions and the corresponding data changes of the user-selected PCA component.

\begin{figure}[h]
    \centering
    \includegraphics[width=1\linewidth]{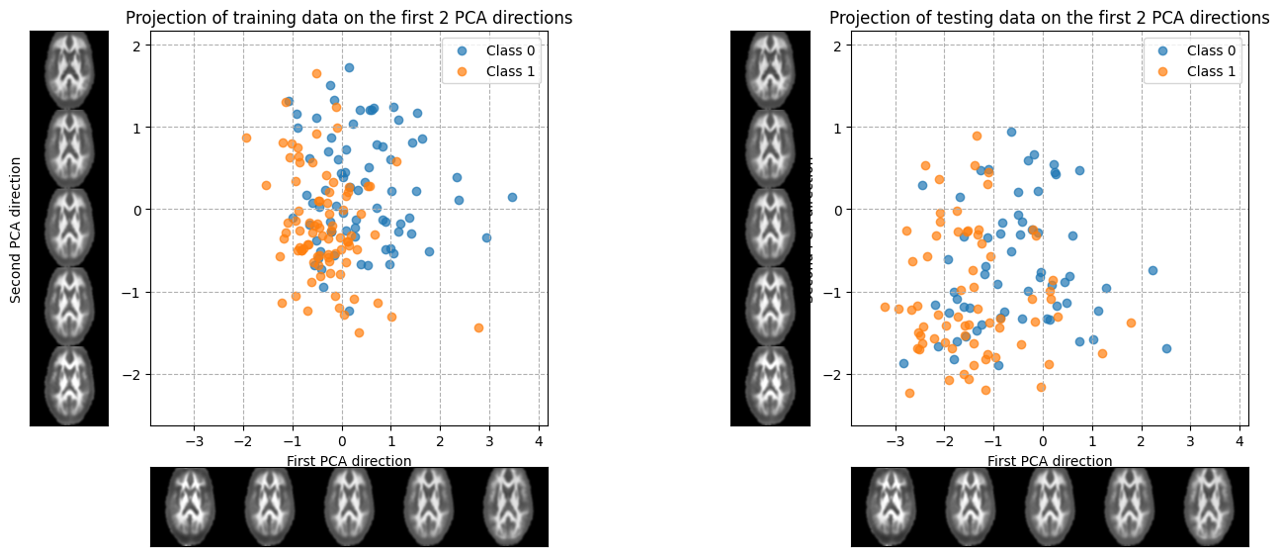}
    \caption{PCA visualization example: distributions of training and testing samples along the projection direction, and the corresponding reconstructed images via inverse transform.}
    \label{fig:pcaexp}
\end{figure}

\begin{code}[h]
\caption{visualize the results of PCA, the code will show the distributions of training and testing samples along
the projection direction, as well the reconstructed images in each PCA mode.(Figure \ref{fig:pcamode} \& Figure \ref{fig:pcaexp})}
\label{code:pca_viusal}
\begin{lstlisting}[language=Python]
from pytranskit.TBM3D.visualization import *
# mode: corresponding component number, from 0 to $pca.n_componets_-1$
visualize_pca(X_train_pca,X_test_pca,train_labels,test_labels,
            reference, pca, modes = [0,1,2,3])
\end{lstlisting}
\end{code}

As shown in Figure \ref{fig:pcaexp}, we visualize the PCA projection of the IXI data set, including the distributions of training and testing samples and their reconstructed images obtained through inverse transform.  These results help illustrate the underlying structure and variability captured in the data set.


\subsubsection{Discriminant analysis}
Discriminant analysis can be employed for data classification. In this work, we adopt the Penalized Linear Discriminant Analysis (PLDA) approach \cite{wang2011penalized}, which aims to obtain a linear mapping of the data such that each image is projected onto a point in a linear subspace. Let $x_i \in R^d$ denote the feature representation of $I_i$ obtained through PCA, and assume that it belongs to class $c$, $\bar{x} = \frac{1}{N} \sum_{i=1}^N x_i$ is the center of the entire data set, $\mu_c = \frac{1}{N_c}\sum_{i\in c} (x_i) $ is the mean of data in class $c$, PLDA solves the following optimization problem:

\begin{equation}
\label{eq:plda}
    \max_w{\frac{w^TS_Tw}{w^T(S_W + \alpha I)w}}
\end{equation}
where $S_T = \sum_{i=1}^N (x_i-\bar{x})(x_i-\bar{x})^T$, $S_W = \sum_c \sum_{i\in c} (x_i-\mu_c)(x_i-\mu_c)^T$, $\alpha$ is a parameter. In Equation (\ref{eq:plda}), the parameter $\alpha$ controls the trade-off between LDA and PCA: lower values emphasize LDA, whereas higher values emphasize PCA \cite{wang2011penalized}.
Then classification is performed based on the mapped representations:

\begin{equation}
    w^Tx + b \gtrless 0
\end{equation}
where $b$ is a threshold.

\begin{code}[h]
\caption{The code for using PLDA for training and prediction.}
\label{code:plda}
\begin{lstlisting}[language=Python]
from pytranskit.optrans.decomposition import PLDA
from pytranskit.TBM3D.gradient import calculate_alpha

#training PLDA
alpha = calculate_alpha(X_train_pca, pca.components_.T, train_labels)
plda=PLDA(alpha=alpha)
plda_proj_tr = plda.fit_transform(X_train_pca,train_labels)[:,0]
plda_proj_te = plda.transform(X_test_pca)[:,0]
\end{lstlisting}
\end{code}

The code \ref{code:plda} provides an example of running PLDA, where the function $calculate\_alpha()$ automatically adjusts an appropriate value of the parameter $\alpha$. This ensures that small variations in $\alpha$ do not lead to substantial changes in the mapping. All the data will be classified based on this mapping. To evaluate the classification performance, 3D-TBM offers multiple metrics, including the confusion matrix and AUROC, as illustrated by the following code.

\begin{code}[h]
\caption{Analysis of the prediction results}
\label{code:plda_results}
\begin{lstlisting}[language=Python]
from pytranskit.TBM3D.visualization import *
from sklearn.metrics import confusion_matrix, ConfusionMatrixDisplay, 
        accuracy_score, balanced_accuracy_score
import matplotlib.pyplot as plt

#preditction
predict_scores = plda.predict(X_test_pca)
acc = accuracy_score(test_labels, predict_scores)
bal_acc = balanced_accuracy_score(test_labels, predict_scores)
cm = confusion_matrix(test_labels, predict_scores)

#show classification results
disp = ConfusionMatrixDisplay(confusion_matrix=cm, display_labels=[0, 1])
disp.plot(cmap='Blues')
plt.title(f'ConfusionMatrix\nAccuracy={acc:.2f},BalancedAccuracy={bal_acc:.2f}')
plt.show()
visual_roc(plda_proj_tr,train_labels,plda_proj_te,test_labels)
\end{lstlisting}
\end{code}


To gain insight into the basis of the classifier’s decisions, we can sample along the discriminant directions and subsequently map these sampled points back to the 3D image domain via the inverse optimal transport maps, according to Equation (\ref{eq:pldavisual}):

\begin{equation}
    \begin{aligned}
        I_{PLDA}(\mathbf{x})=(D_{f_\alpha^{-1}}(\mathbf{x})'I_0( f_\alpha^{-1}(\mathbf{x})),\\  f_\alpha = \bar{f}+\alpha w^TB,\  \alpha = -2\sigma,-1\sigma,0,\sigma,2\sigma
    \end{aligned}
\label{eq:pldavisual}     
\end{equation}
where $\sigma$ is the standard deviation of the data along the discriminant direction, $w$ is obtained by Equation (\ref{eq:plda}) and $B$ is obtained by PCA. By observing the resulting morphological variations, we can identify the most discriminative structural features. Code \ref{code:inverse_viusal} visualizes the distribution of data from each class along the discriminant direction and presents the corresponding 3D images.

\begin{figure}[!h]
    \centering
    \includegraphics[width=0.8\linewidth]{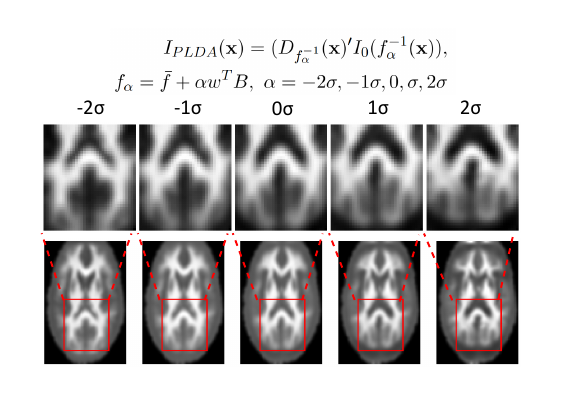}
    \caption{Visualization of PLDA in IXI data set}
    \label{fig:pldaexp}
\end{figure}

\begin{code}[h]
\caption{Visualize the PLDA results by inverse transform.(Figure \ref{fig:pldaexp})}
\label{code:inverse_viusal}
\begin{lstlisting}[language=Python]
#visualization PLDA
inverse_visualiztion(plda_proj_tr,train_labels,reference,plda,pca)
\end{lstlisting}
\end{code}

We provide an interface for visualizing classification results, which also fits the data distribution along the discriminant direction. Figure \ref{fig:pldaexp} shows the visualization results of age classification based on white matter images from the IXI dataset, where samples located further to the left tend to correspond to younger patients, whereas those on the right are more likely to belong to older patients. From the visualization, we can observe that he ventricle size is increasing with age, which is consistent with previous findings\cite{kundu2021investigating}.


\subsubsection{Canonical correlation analysis}

If the user wishes to perform correlation analysis or regression, 3D-TBM provides tools for canonical correlation analysis (CCA) and linear regression \cite{hardoon2004canonical}. Suppose $X\in R^{n*d}$ represents the feature representations obtained after PCA, and $Y\in R^{n*q}$  denotes another set of variables for analysis, such as age($q=1$ in this case). Let $\Sigma_{XX}$ and $\Sigma_{YY}$ be the covariance matrices of $X$ and $y$, respectively, and $\Sigma_{XY}$ be their cross-covariance matrix. Then, CCA aims to solve the following optimization problem:

\begin{equation}
\begin{aligned}
\max_{a, b} \quad & \rho = \frac{a^\top \Sigma_{XY} b}{\sqrt{a^\top \Sigma_{XX} a }\;\sqrt{ b^\top \Sigma_{YY} b}} \\
\text{s.t.} \quad & a^\top \Sigma_{XX} a = 1, \quad b^\top \Sigma_{YY} b = 1
\end{aligned}
\end{equation}

Subsequently, $X$ is projected onto the canonical direction $a$, and the resulting projection is regressed against the associated variable for further analysis. The implementation is illustrated in the code \ref{code:cca}:

\begin{code}[h]
\caption{Training and testing by CCA}
\label{code:cca}
\begin{lstlisting}[language=Python]
import numpy as np
from pytranskit.optrans.decomposition import CCA
from sklearn.linear_model import LinearRegression

#canonical correlation analysis
cca = CCA()
cca_proj_tr, cca_proj_tr_label = cca.fit_transform(X_train_pca,train_labels)
cca_proj_te = cca.transform(X_test_pca)

#linear regression
reg = LinearRegression()
reg.fit(cca_proj_tr, train_labels)
Y_pred = reg.predict(cca_proj_tr)
\end{lstlisting}
\end{code}

Similar to PLDA, we also visualize the regression results by sampling along the projection direction and performing inverse transforms to map them back to the image domain, enabling the observation of how image-domain features relate to other variables(code \ref{code:inverse_cca}).

\begin{figure}[!h]
    \centering
    \includegraphics[width=1\linewidth]{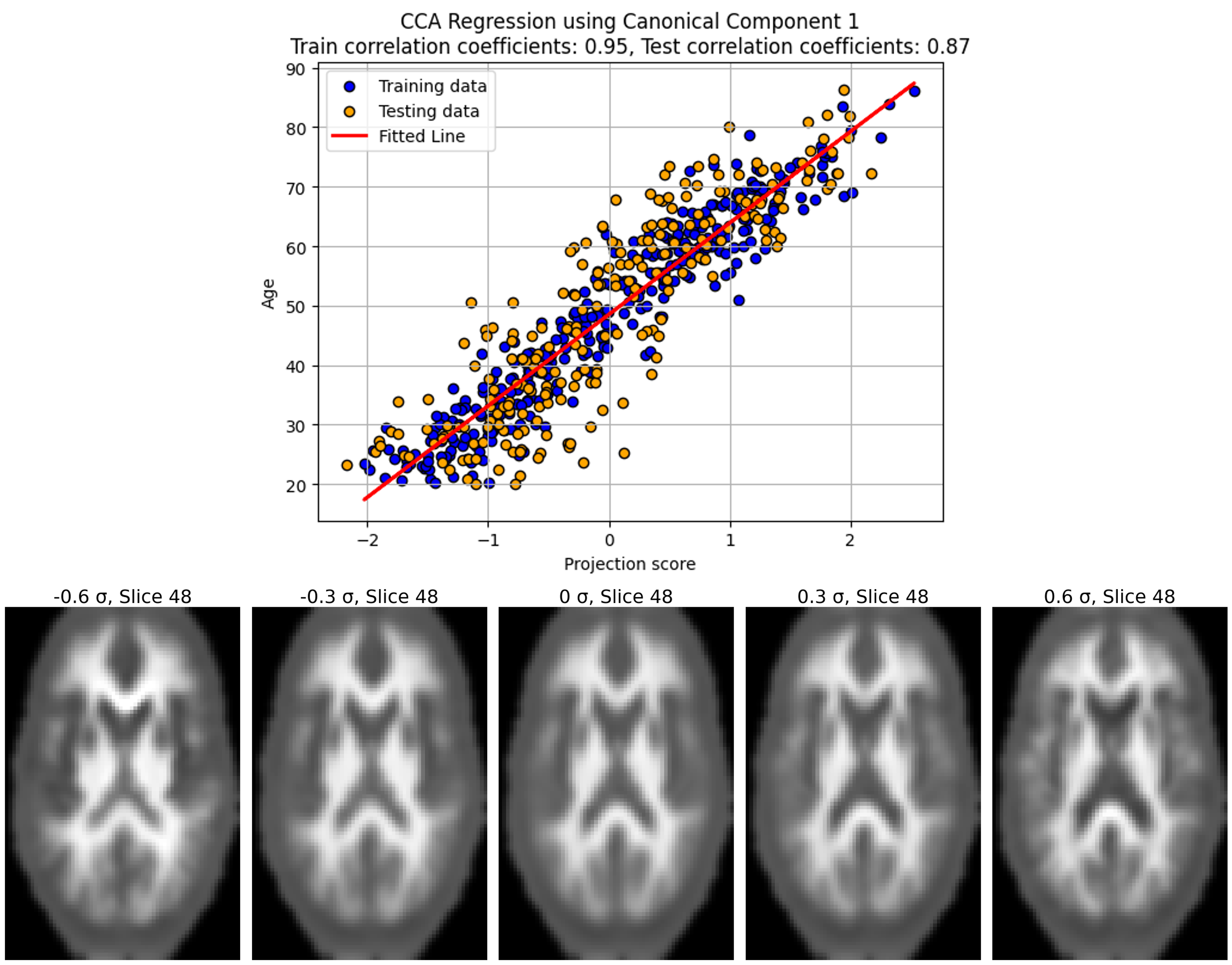}
    \caption{Visualization of CCA in IXI data set}
    \label{fig:ccaexp}
\end{figure}

\begin{code}[h]
\caption{Visualize the CCA results by inverse transform.(Figure \ref{fig:ccaexp})}
\label{code:inverse_cca}
\begin{lstlisting}[language=Python]
from pytranskit.TBM3D.visualization import *

#Set sampling interval
data_std = np.std(cca_proj_tr)
vec = np.zeros((5,cca.n_components_))
stds = [-1,-0.5,0,0.5,1]
vec[:,0] = np.asarray(stds)*data_std

#Inverse transform
cca.mean_ = 0
inverse_pca = cca.inverse_transform(vec)
X_inverse = pca.inverse_transform(inverse_pca)
images = inverse_image(X_inverse, reference)
show_3d_arrays(images,stds=stds)
\end{lstlisting}
\end{code}

The results of the age analysis based on the white matter regions are shown in Figure \ref{fig:ccaexp}. From the visualization, we obtained conclusions similar to those of PLDA, namely that ventricular size increases with age.


\subsubsection{Classification}

In this section, we introduce several additional classifiers suitable for TBM-based analysis, such as Nearest Subspace\cite{shifat2021radon} and Local Nearest Subspace\cite{rubaiyat2024end}, and design experiments to compare their performance. This evaluation highlights the improvements in predictive accuracy afforded by these classification methods.

We conducted experiments on the IXI dataset, aiming to classify subjects into young or old based on their white‐matter MRI volumes. A total of 330 samples were included and randomly split into a training set and a test set with a 200/130 ratio. To reduce the influence of random variation, this random split was repeated 20 times, and a new classifier was trained for each split.

\begin{table}[htbp]
\centering
\begin{tabular}{c|cc}
\hline
Classifier& ACC & Balanced acc\\
\hline
PLDA & 0.78±0.026 & 0.777±0.028 \\
Nearest Subspace & 0.846±0.033 & 0.84±0.034 \\
Local Nearest Subspace & 0.82±0.042 & 0.822±0.041 \\
\hline
\end{tabular}
\caption{Results on classification}
\label{tab:exp}
\end{table}

The experimental results are summarized in Table \ref{tab:exp}. We observe that, compared with the linear classifier (PLDA), both the Nearest Subspace classifier shows a noticeable improvement(Acc from 0.78 to 0.846), suggesting that this classification task may better satisfy its underlying assumptions. Meanwhile, the Local Nearest Subspace performs less favorably, which may indicate that the task is relatively simple and that the data distribution in the transport domain does not exhibit significant complexity.

Beyond these classifiers, users can leverage other analysis tools from the scikit-learn library \cite{pedregosa2011scikit}, such as support vector machines (SVM) \cite{cortes1995support}, k-nearest neighbors (kNN) \cite{cover1967nearest}, random forests (RF) \cite{breiman2001random}, deep learning\cite{lecun2002gradient}, or implement their own custom algorithms to suit specific objectives.





\section{Summary}

We present 3D-TBM, a tool that applies transport-based morphometry (TBM) to 3D medical imaging. This paper introduces the motivation behind the design of 3D-TBM and its overall workflow, and it provides a detailed description of each functional component and its usage. Moreover, we include illustrative examples to facilitate ease of use, with the aim of enabling researchers to readily incorporate TBM into their clinical studies. Finally, we designed a simple experiment to compare classification performance under different choices of references and classifiers, and the results indicate that employing the intrinsic mean as the reference may improve the predictive accuracy of the classifiers.



\acks{Authors gratefully acknowledge funding from the ONR (N000142212505) and the NIH (GM130825, U54-CA274499) in contributing to a portion of this work. 
Authors also acknowledge the source of the IXI data: https://brain-development.org/ixi-dataset/ .}











\vskip 0.2in
\bibliographystyle{IEEEtran}
\bibliography{references}

\begin{thebibliography}{10}
\providecommand{\url}[1]{#1}
\csname url@samestyle\endcsname
\providecommand{\newblock}{\relax}
\providecommand{\bibinfo}[2]{#2}
\providecommand{\BIBentrySTDinterwordspacing}{\spaceskip=0pt\relax}
\providecommand{\BIBentryALTinterwordstretchfactor}{4}
\providecommand{\BIBentryALTinterwordspacing}{\spaceskip=\fontdimen2\font plus
\BIBentryALTinterwordstretchfactor\fontdimen3\font minus \fontdimen4\font\relax}
\providecommand{\BIBforeignlanguage}[2]{{%
\expandafter\ifx\csname l@#1\endcsname\relax
\typeout{** WARNING: IEEEtran.bst: No hyphenation pattern has been}%
\typeout{** loaded for the language `#1'. Using the pattern for}%
\typeout{** the default language instead.}%
\else
\language=\csname l@#1\endcsname
\fi
#2}}
\providecommand{\BIBdecl}{\relax}
\BIBdecl

\bibitem{pytranskit2023}
A.~H.~M. Rubaiyat, X.~Yin, L.~Cattell, S.~Kolouri, M.~Shifat-E-Rabbi, Y.~Zhuang, and G.~Rohde, ``Pytranskit: Python transport based signal processing toolkit,'' \url{https://github.com/rohdelab/PyTransKit}, 2023, gitHub repository.

\bibitem{kundu2018discovery}
S.~Kundu, S.~Kolouri, K.~I. Erickson, A.~F. Kramer, E.~McAuley, and G.~K. Rohde, ``Discovery and visualization of structural biomarkers from mri using transport-based morphometry,'' \emph{NeuroImage}, vol. 167, pp. 256--275, 2018.

\bibitem{wang2013linear}
W.~Wang, D.~Slep{\v{c}}ev, S.~Basu, J.~A. Ozolek, and G.~K. Rohde, ``A linear optimal transportation framework for quantifying and visualizing variations in sets of images,'' \emph{International journal of computer vision}, vol. 101, no.~2, pp. 254--269, 2013.

\bibitem{ironside2024fully}
N.~Ironside, VISTA-ICH, K.~El~Naamani, T.~Rizvi, M.~S. El-Rabbi, S.~Kundu, A.~Becceril-Gaitan, C.-J. Chen, S.~A. Mayer, E.~S. Connolly \emph{et~al.}, ``Fully automated hematoma expansion prediction from non-contrast computed tomography in spontaneous ich patients,'' \emph{medRxiv}, pp. 2024--05, 2024.

\bibitem{kundu2021investigating}
S.~Kundu, H.~Huang, K.~I. Erickson, E.~McAuley, A.~F. Kramer, and G.~K. Rohde, ``Investigating impact of cardiorespiratory fitness in reducing brain tissue loss caused by ageing,'' \emph{Brain Communications}, vol.~3, no.~4, p. fcab228, 2021.

\bibitem{kundu2024discovering}
S.~Kundu, H.~Sair, E.~H. Sherr, P.~Mukherjee, and G.~K. Rohde, ``Discovering the gene-brain-behavior link in autism via generative machine learning,'' \emph{Science advances}, vol.~10, no.~24, p. eadl5307, 2024.

\bibitem{villani2008optimal}
C.~Villani \emph{et~al.}, \emph{Optimal transport: old and new}.\hskip 1em plus 0.5em minus 0.4em\relax Springer, 2008, vol. 338.

\bibitem{kolouri2017optimal}
S.~Kolouri, S.~R. Park, M.~Thorpe, D.~Slepcev, and G.~K. Rohde, ``Optimal mass transport: Signal processing and machine-learning applications,'' \emph{IEEE signal processing magazine}, vol.~34, no.~4, pp. 43--59, 2017.

\bibitem{basu2014detecting}
S.~Basu, S.~Kolouri, and G.~K. Rohde, ``Detecting and visualizing cell phenotype differences from microscopy images using transport-based morphometry,'' \emph{Proceedings of the National Academy of Sciences}, vol. 111, no.~9, pp. 3448--3453, 2014.

\bibitem{jolliffe2016principal}
I.~T. Jolliffe and J.~Cadima, ``Principal component analysis: a review and recent developments,'' \emph{Philosophical transactions of the royal society A: Mathematical, Physical and Engineering Sciences}, vol. 374, no. 2065, p. 20150202, 2016.

\bibitem{balakrishnama1998linear}
S.~Balakrishnama and A.~Ganapathiraju, ``Linear discriminant analysis-a brief tutorial,'' \emph{Institute for Signal and information Processing}, vol.~18, no. 1998, pp. 1--8, 1998.

\bibitem{wang2011penalized}
W.~Wang, Y.~Mo, J.~A. Ozolek, and G.~K. Rohde, ``Penalized fisher discriminant analysis and its application to image-based morphometry,'' \emph{Pattern recognition letters}, vol.~32, no.~15, pp. 2128--2135, 2011.

\bibitem{hardoon2004canonical}
D.~R. Hardoon, S.~Szedmak, and J.~Shawe-Taylor, ``Canonical correlation analysis: An overview with application to learning methods,'' \emph{Neural computation}, vol.~16, no.~12, pp. 2639--2664, 2004.

\bibitem{avants2009advanced}
B.~B. Avants, N.~Tustison, G.~Song \emph{et~al.}, ``Advanced normalization tools (ants),'' \emph{Insight j}, vol.~2, no. 365, pp. 1--35, 2009.

\bibitem{tustison2021antsx}
N.~J. Tustison, P.~A. Cook, A.~J. Holbrook, H.~J. Johnson, J.~Muschelli, G.~A. Devenyi, J.~T. Duda, S.~R. Das, N.~C. Cullen, D.~L. Gillen \emph{et~al.}, ``The antsx ecosystem for quantitative biological and medical imaging,'' \emph{Scientific reports}, vol.~11, no.~1, p. 9068, 2021.

\bibitem{tustison2024antsx}
N.~J. Tustison, M.~A. Yassa, B.~Rizvi, P.~A. Cook, A.~J. Holbrook, M.~T. Sathishkumar, M.~G. Tustison, J.~C. Gee, J.~R. Stone, and B.~B. Avants, ``Antsx neuroimaging-derived structural phenotypes of uk biobank,'' \emph{Scientific Reports}, vol.~14, no.~1, p. 8848, 2024.

\bibitem{cardoso2022monai}
M.~J. Cardoso, W.~Li, R.~Brown, N.~Ma, E.~Kerfoot, Y.~Wang, B.~Murrey, A.~Myronenko, C.~Zhao, D.~Yang \emph{et~al.}, ``Monai: An open-source framework for deep learning in healthcare,'' \emph{arXiv preprint arXiv:2211.02701}, 2022.

\bibitem{fischl2012freesurfer}
B.~Fischl, ``Freesurfer,'' \emph{Neuroimage}, vol.~62, no.~2, pp. 774--781, 2012.

\bibitem{dale1999cortical}
A.~M. Dale, B.~Fischl, and M.~I. Sereno, ``Cortical surface-based analysis: I. segmentation and surface reconstruction,'' \emph{Neuroimage}, vol.~9, no.~2, pp. 179--194, 1999.

\bibitem{friston1994statistical}
K.~J. Friston, A.~P. Holmes, K.~J. Worsley, J.-P. Poline, C.~D. Frith, and R.~S. Frackowiak, ``Statistical parametric maps in functional imaging: a general linear approach,'' \emph{Human brain mapping}, vol.~2, no.~4, pp. 189--210, 1994.

\bibitem{penny2011statistical}
W.~D. Penny, K.~J. Friston, J.~T. Ashburner, S.~J. Kiebel, and T.~E. Nichols, \emph{Statistical parametric mapping: the analysis of functional brain images}.\hskip 1em plus 0.5em minus 0.4em\relax Elsevier, 2011.

\bibitem{cox1996afni}
R.~W. Cox, ``Afni: software for analysis and visualization of functional magnetic resonance neuroimages,'' \emph{Computers and Biomedical research}, vol.~29, no.~3, pp. 162--173, 1996.

\bibitem{cox1997software}
R.~W. Cox and J.~S. Hyde, ``Software tools for analysis and visualization of fmri data,'' \emph{NMR in Biomedicine: An International Journal Devoted to the Development and Application of Magnetic Resonance In Vivo}, vol.~10, no. 4-5, pp. 171--178, 1997.

\bibitem{flamary2021pot}
R.~Flamary, N.~Courty, A.~Gramfort, M.~Z. Alaya, A.~Boisbunon, S.~Chambon, L.~Chapel, A.~Corenflos, K.~Fatras, N.~Fournier \emph{et~al.}, ``Pot: Python optimal transport,'' \emph{Journal of Machine Learning Research}, vol.~22, no.~78, pp. 1--8, 2021.

\bibitem{ixi_dataset}
``Ixi – information extraction from images (brain-development.org/ixi-dataset),'' \url{https://brain-development.org/ixi-dataset/}, 2003, accessed: 2026-01-14.

\bibitem{harris2020array}
C.~R. Harris, K.~J. Millman, S.~J. Van Der~Walt, R.~Gommers, P.~Virtanen, D.~Cournapeau, E.~Wieser, J.~Taylor, S.~Berg, N.~J. Smith \emph{et~al.}, ``Array programming with numpy,'' \emph{nature}, vol. 585, no. 7825, pp. 357--362, 2020.

\bibitem{kolouri2016continuous}
S.~Kolouri, A.~B. Tosun, J.~A. Ozolek, and G.~K. Rohde, ``A continuous linear optimal transport approach for pattern analysis in image datasets,'' \emph{Pattern recognition}, vol.~51, pp. 453--462, 2016.

\bibitem{agueh2011barycenters}
M.~Agueh and G.~Carlier, ``Barycenters in the wasserstein space,'' \emph{SIAM Journal on Mathematical Analysis}, vol.~43, no.~2, pp. 904--924, 2011.

\bibitem{shifat2021radon}
M.~Shifat-E-Rabbi, X.~Yin, A.~H.~M. Rubaiyat, S.~Li, S.~Kolouri, A.~Aldroubi, J.~M. Nichols, and G.~K. Rohde, ``Radon cumulative distribution transform subspace modeling for image classification,'' \emph{Journal of Mathematical Imaging and Vision}, vol.~63, no.~9, pp. 1185--1203, 2021.

\bibitem{rubaiyat2024end}
A.~H.~M. Rubaiyat, S.~Li, X.~Yin, M.~Shifat-E-Rabbi, Y.~Zhuang, and G.~K. Rohde, ``End-to-end signal classification in signed cumulative distribution transform space,'' \emph{IEEE Transactions on Pattern Analysis and Machine Intelligence}, vol.~46, no.~9, pp. 5936--5950, 2024.

\bibitem{pedregosa2011scikit}
F.~Pedregosa, G.~Varoquaux, A.~Gramfort, V.~Michel, B.~Thirion, O.~Grisel, M.~Blondel, P.~Prettenhofer, R.~Weiss, V.~Dubourg \emph{et~al.}, ``Scikit-learn: Machine learning in python,'' \emph{the Journal of machine Learning research}, vol.~12, pp. 2825--2830, 2011.

\bibitem{cortes1995support}
C.~Cortes and V.~Vapnik, ``Support-vector networks,'' \emph{Machine learning}, vol.~20, no.~3, pp. 273--297, 1995.

\bibitem{cover1967nearest}
T.~Cover and P.~Hart, ``Nearest neighbor pattern classification,'' \emph{IEEE transactions on information theory}, vol.~13, no.~1, pp. 21--27, 1967.

\bibitem{breiman2001random}
L.~Breiman, ``Random forests,'' \emph{Machine learning}, vol.~45, no.~1, pp. 5--32, 2001.

\bibitem{lecun2002gradient}
Y.~LeCun, L.~Bottou, Y.~Bengio, and P.~Haffner, ``Gradient-based learning applied to document recognition,'' \emph{Proceedings of the IEEE}, vol.~86, no.~11, pp. 2278--2324, 2002.

\end{thebibliography}

\end{document}